\documentclass[letterpaper, 10 pt, conference]{ieeeconf}  

\usepackage{amsmath,amsfonts}

\usepackage{array}
\usepackage{float}
\usepackage{algorithm}
\usepackage{algpseudocode}
\usepackage[hidelinks]{hyperref}
\usepackage{textcomp}
\usepackage{stfloats}
\usepackage{graphicx}
\usepackage{subcaption}
\usepackage{subcaption}
\usepackage{bm}
\usepackage{pifont}
\usepackage{hyperref}
\usepackage{booktabs} 
\hypersetup{
	colorlinks=true, 
	linkcolor=blue,  
	citecolor=blue, 
	urlcolor=red     
}
\usepackage{url}
\usepackage{verbatim}
\usepackage{graphicx}
\usepackage{xcolor}
\def\BibTeX{{\rm B\kern-.05em{\sc i\kern-.025em b}\kern-.08em
		T\kern-.1667em\lower.7ex\hbox{E}\kern-.125emX}}
\usepackage{balance}

\IEEEoverridecommandlockouts        

\usepackage{amsmath}

\title{\LARGE \bf

Whole-Body Semantic-to-Actuation Grounding of Elephant-Inspired Soft-Trunk Motion via Lightweight Flow Matching

}

\author{Tingcong Liu$^{1,2}$, Tongshun Chen$^{3}$, Siyi Ma$^{4}$, Yuhao Wang$^{3}$, Aye Phyu Phyu Aung$^{2}$, Ibrahim Alsarraj$^{3}$,\\
J. Senthilnath$^{2}$, Bo An$^{1}$, Ke Wu$^{3}$
}

\begin{document}

\maketitle
\thispagestyle{empty}
\pagestyle{empty}

\setlength{\textfloatsep}{5pt}    
\setlength{\intextsep}{5pt}       
\begin{abstract}
For close-contact human--robot interaction (HRI), trunk-like continuum
manipulators offer an attractive physical channel for diverse whole-body
shape-based expression. However, directly applying conventional
vision--language--action (VLA) grounding to such systems is challenging:
tip- or end-effector motion underspecifies whole-body expression, whereas
direct whole-body specification is too high-dimensional for semantic planning
and difficult to keep feasible. To address this representation mismatch, we
propose a whole-body semantic-to-actuation grounding framework for
elephant-inspired soft-trunk HRI via lightweight flow matching. The framework
grounds open-vocabulary responses produced by a multimodal large language
model (MLLM) into executable tendon-actuation trajectories that realize
distributed continuum-body motion. It first canonicalizes open-vocabulary
responses into bounded morphology-aligned intent--intensity tuples that
define executable motion families, and then parameterizes continuous
whole-body tendon-actuation profiles with compact Catmull--Rom spline
controls. Conditioned on the resulting tuple, a lightweight rectified-flow
generator learns a distribution over feasible spline controls, enabling
efficient one-to-many sampling of smooth and socially valid trunk-motion
realizations. We evaluate the framework through semantic-to-actuation
grounding ablations and a 100-participant physical HRI study. In the
experiments, the proposed framework improves held-out grounding correctness
from $25.0\%$ to $77.2\%$ over a raw-response dense-regression baseline.
Compared with a denoising-diffusion baseline, rectified flow improves
correctness from $71.9\%$ to $77.2\%$ and reduces inference time from
$7.86$ ms to $4.87$ ms, while preserving motion diversity. In the physical HRI study, the proposed framework increases the positive
overall-satisfaction rating from $46\%$ to $82\%$ over the audiovisual-only
baseline, demonstrating the user-facing benefit of the generated soft-trunk
motion channel.
\end{abstract}


 \section{INTRODUCTION}

Close-contact human--robot interaction (HRI) brings the robot body into
the user's interpersonal and physical space, where approach distance,
bodily behavior, and touch shape perceived appropriateness, comfort,
intent, and safety~\cite{mumm2011proxemics,shiomi2020survey}. Since
robot motion can function as a communicative resource beyond speech or
screen-based display~\cite{hoffman2014designing}, close-contact HRI
motivates a physical extension that can act as an embodied channel for
social response. Rigid manipulators offer accurate and repeatable
motion, but close-contact social use must also address physical safety,
perceived safety, and comfort in proximity~\cite{hamad2023concise}.
Trunk-like soft continuum manipulators are therefore promising: their
continuum bodies can generate non-contact body language through
approach, withdrawal, and gesture~\cite{cha2018survey}, while their
compliant morphology can deform during light contact~\cite{rus2015design}.
However, the same high-dimensional and deformable whole-body morphology
that enables expressive and contact-friendly interaction also changes
the grounding problem: the behavior perceived by the user is distributed
over the entire continuum body rather than concentrated at a single end
effector. 
This creates a mismatch with the tip-centered action representations
commonly used in vision-language-action robot
policies~\cite{brohan2022rt}. Existing soft-robot VLA-style
formulations~\cite{wei2026manisoft} typically represent robot actions
through end-effector displacements or low-level control commands. Such
representations are suitable for manipulation tasks that demand precise
task-space outcomes, but are insufficient for soft trunks whose social
behavior emerges from whole-body
deformation~\cite{webster2010design}. 
Conversely, directly predicting body-shape over-specifies the response and creates a high-dimensional
learning target that is difficult to learn and keep
feasible~\cite{della2023model}. Motivated by this observation, we depart from the conventional VLA paradigm of grounding language directly into task-space actions. Instead, we seek to ground open-vocabulary semantic responses produced by MLLMs into executable whole-body trunk actuation trajectories. In this context, a whole-body semantic-to-actuation grounding interface provides a feasible way to bridge semantic responses and morphology-aligned motion.

However, such an interface also faces some challenges: Open-vocabulary MLLM responses may contain unsupported actions/motions,
ambiguous modifiers, or non-physical expressions that are not directly
executable by the trunk morphology~\cite{ahn2022can}, while dense
actuation trajectories are weakly structured and difficult to learn
from limited demonstrations~\cite{ijspeert2013dynamical}. Beyond executability, the same social
behavior may admit multiple feasible realizations~\cite{seker2019conditional}, so the grounding
model should capture a conditional distribution over executable motions
rather than collapse each semantic condition to a single deterministic
trajectory~\cite{florence2022implicit}. Diffusion-based generators~\cite{ho2020denoising}  are natural candidates for such
distributional motion generation, but their iterative denoising process
can be costly for responsive close-contact HRI.
Rectified flow~\cite{liu2022flow} is a lightweight flow-matching
alternative: it learns a continuous transport field from noise to the limited expert motion manifold, retaining distributional generation with a
simpler sampling process~\cite{lipman2023flow}. Recent flow-matching applications in soft robotics, including inverse dynamics~\cite{yang2026flow} and task-level soft grasping~\cite{yang2026lightweight}, demonstrate its promise for efficient generation of diverse actuation trajectories. However, its use for semantic-to-actuation grounding in long-horizon expressive HRI remains unexplored. The key idea is to treat semantic grounding for soft-trunk HRI as a morphology-aligned whole-body actuation representation problem,
rather than as a conventional language-to-tip-action mapping problem. {Motivated by this, the key contributions of the paper are proposed and summarized as follows:
\begin{figure*}[t]
   \centering
\includegraphics[page=1,width=178mm,trim=0mm 20mm 0mm 40mm,clip]{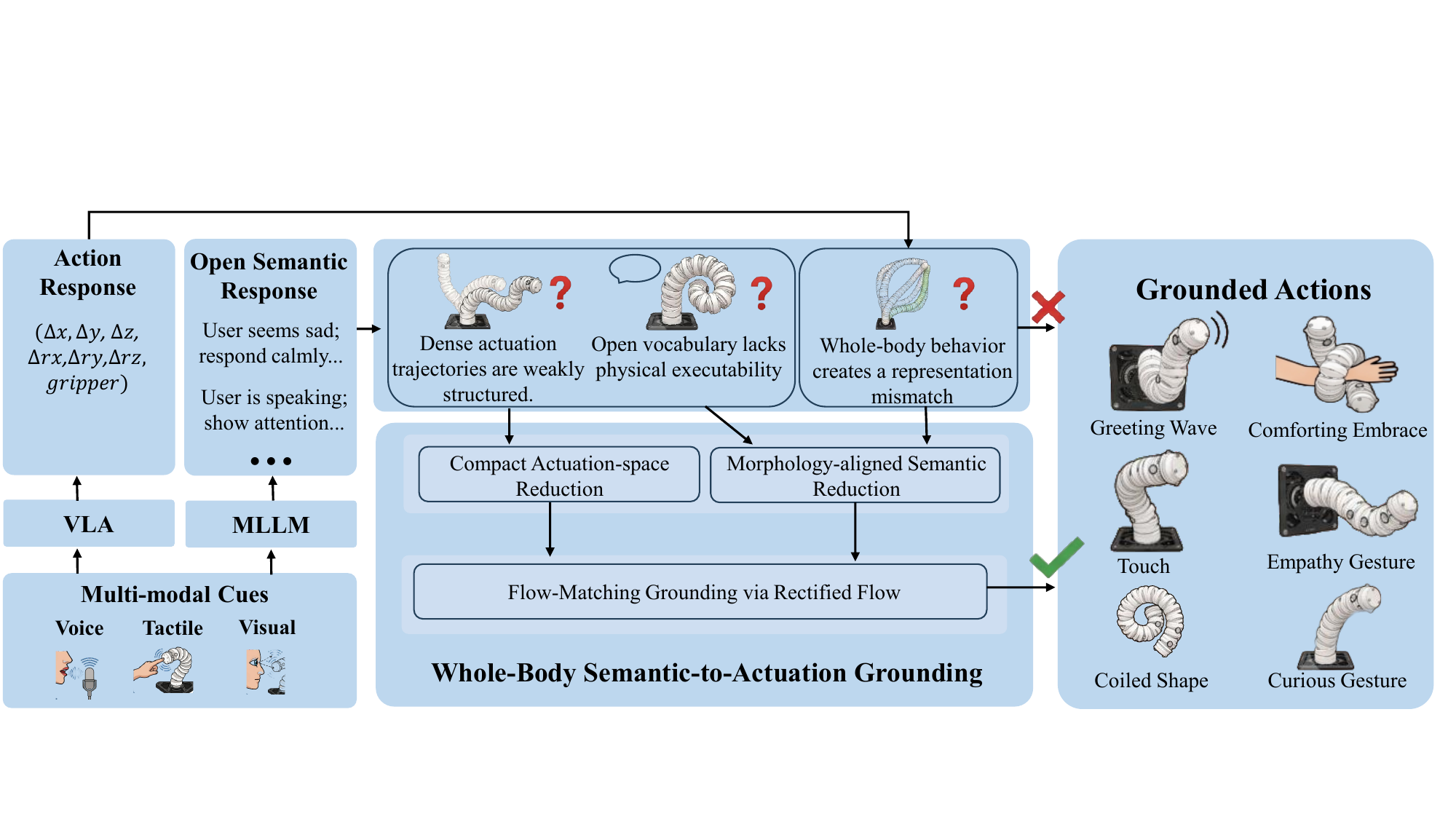}
   \caption{Overview of the whole-body semantic-to-actuation grounding framework.}
   \label{fig:overview}
\end{figure*}
\begin{enumerate}
    \item \textbf{Whole-body semantic-to-actuation grounding:} We propose a whole-body semantic-to-actuation grounding framework for close-contact soft-trunk HRI, where open semantic responses from an MLLM are grounded as executable whole-body soft-trunk motions.
    \item \textbf{Dual reduction for trunk-motion representations:}
    We introduce a dual-reduction representation that canonicalizes
    open-vocabulary multimodal responses into bounded
    morphology-aligned intent--intensity tuples and parameterizes
    continuous whole-body tendon-actuation profiles with compact
    Catmull--Rom spline controls.
  \item \textbf{Lightweight flow-matching learning:}
    We develop a lightweight tuple-conditioned rectified-flow generator
    that learns a distribution over reduced spline controls, enabling one-to-many sampling of smooth and socially
    valid trunk-motion realizations with efficient inference.
    \item \textbf{Ablation experiment and physical HRI study:}
We validate the pipeline through ablation experiments and a 100-participant physical HRI study. The proposed framework improves held-out correctness from 25.0\% to 77.2\% over a raw-response dense-regression baseline, outperforms DDPM in both correctness and inference speed, and increases the positive overall-satisfaction rating from 46\% to 82\% compared with the baseline setup.
\end{enumerate}}

\section{Problem Statement}

{
We study whole-body semantic-to-actuation grounding for close-contact soft-trunk HRI. Given an open-vocabulary response produced by
an MLLM, the goal is to ground it into a morphology-compatible
soft-trunk motion family and realize it as a socially appropriate,
expressive, and executable tendon-actuation trajectory. This leads to three challenges (Fig.~\ref{fig:overview}):
}



\begin{itemize}
    \item \textbf{Whole-body behavior creates a representation mismatch.}
    Tip-level trajectories underspecify body-shape trunk behavior, whereas whole-body trajectories overspecify it and create a
    high-dimensional learning target.
    {
    \item \textbf{Dense actuation trajectories are weakly
structured.}
Directly predicting a dense whole-body actuation trajectory from a
semantic condition forces the model to learn hundreds of weakly
structured values.
    }
    \item \textbf{Open vocabulary lacks physical executability.}
    Open responses may contain unsupported actions/motions, ambiguous
    modifiers, or non-physical expressions, and may cause physically infeasible and socially drifting motions.
\end{itemize}

\section{Methodology}
The framework consists of three components, as shown in Fig.~\ref{fig:framework}. First, morphology-aligned semantic reduction canonicalizes open-vocabulary multimodal responses into bounded intent--intensity tuples. Second, compact actuation-space reduction parameterizes dense whole-body actuation trajectories as low-dimensional spline-control matrices. Third, flow-matching grounding constructs reduced demonstration pairs and learns a conditional distribution over spline controls given the morphology-aligned intent--intensity tuple, enabling efficient sampling of executable whole-body trunk motions.

\subsection{Morphology-aligned Semantic Reduction}
\label{subsec:reduction}
This component reduces open-vocabulary multimodal reasoning 
to an executable semantic condition for soft-trunk grounding. 



\subsubsection{Multimodal Response Reasoning}
At interaction turn \(t\), the interface receives visual, vocal, and trunk-based tactile observation cues,
\begin{equation}
\small
    o_t=(o_t^{\mathrm{vis}},o_t^{\mathrm{voc}},o_t^{\mathrm{tac}}),
\end{equation}
together with a recent history \(\mathcal{H}_{t-1}\). The history stores observation cues $o_{t-1}$, recent trunk intents $m_{t-1}$, and intensity levels $z_{t-1}$ defined in Section \ref{Morp}. Inspired by MLLM-agent pipelines for observation-conditioned reasoning and
decision making~\cite{yao2022react}, the MLLM reasons over the current cues and history to generate an
open-vocabulary natural-language response plan:
\begin{equation}
\label{eq:2}
\small
    r_t=f_{\mathrm{MLLM}}(o_t,\mathcal{H}_{t-1}).
\end{equation}
\(r_t\) is a general HRI response describing the robot's intended social reaction, such
as "User seems sad; respond calmly...".

\subsubsection{Executable Semantic Reduction}
\label{Morp}

As shown in Fig. \ref{fig:framework}(A), although \(r_t\) describes the desired social reaction, it is not
directly executable by the soft trunk. Paraphrases of the same behavior
may be mapped to different language embeddings, whose geometry does not
necessarily reflect similarities between executable motions. Moreover,
inspired by the elephant trunk's use of a compact kinematic repertoire
to manage its high-dimensional morphology~\cite{dagenais2021elephant},
the platform's executable vocabulary is organized around
morphology-specific whole-body patterns, including bending, curling,
extension, retraction, and compliant contact. Direct conditioning on
\(r_t\) would therefore entangle linguistic variation,
morphology-dependent feasibility, and continuous trajectory generation. Accordingly, a morphology-aligned semantic reduction maps \(r_t\) to
a bounded tuple
\((m_t,z_t)\in\mathcal{M}\times\mathcal{Z}\), thereby separating
executable semantic selection from continuous motion realization. The
intended tuple-selection criterion is conceptually formulated as
\begin{equation}
\small
\label{eq:math_loss}
    (m_t,z_t)
    \approx
    \arg\min_{m\in\mathcal{M},\,z\in\mathcal{Z}}
    \mathcal{L}_{\mathrm{sem}}(r_t,d_{m,z};\mathcal{C}),
\end{equation}
where \(d_{m,z}\) is the canonical morphology-compatible description
associated with intent \(m\) at interaction intensity \(z\), and
\(\mathcal{L}_{\mathrm{sem}}\) denotes its semantic mismatch with
\(r_t\) under the grounding specification \(\mathcal{C}\). The intent \(m_t\in\mathcal{M}\) selects one of sixteen executable
whole-body elephant-inspired social-motion classes, including greeting, comforting,
attentive listening, curiosity, embracing, withdrawal, coiling and so on. The
ordered interaction-intensity variable \(z_t\in\mathcal{Z}\) selects
one of four execution profiles: \textit{Subtle}, \textit{Moderate},
\textit{Pronounced}, and \textit{Emphatic}. These levels represent
increasing embodied interaction salience through intent-dependent
changes in reach, curvature, orientation, temporal emphasis, and
contact proximity. Each tuple identifies an executable motion family, leaving the remaining within-tuple realization
variation to the downstream flow model. The grounding specification \(\mathcal{C}\) is instantiated as a
robot-specific few-shot in-context prompt encoding the trunk
affordances and feasibility limits, the supported intent--intensity
space, canonical tuple descriptions, response-to-tuple examples, and a
structured output schema. Eq.~\eqref{eq:math_loss}
defines a conceptual selection criterion. In practice, this reduction process is implemented by the MLLM under the grounding specification \(\mathcal{C}\):
\begin{equation}
\label{eq:4}
\small
    (m_t,z_t)
    =
    f_{\mathrm{MLLM}}(r_t;\mathcal{C}).
\end{equation}
As shown in the morphology-aligned semantic-reduction block of
Fig.~\ref{fig:framework}(A), the resulting tuple \((m_t,z_t)\) provides a
bounded, morphology-compatible semantic condition for the grounding
model.

\subsection{Compact Actuation-space Reduction}
\label{subsec:trunk_grounding}

\begin{figure*}[t]
   \centering
\includegraphics[page=3,width=178mm,trim=0mm 5mm 0mm 40mm,clip]{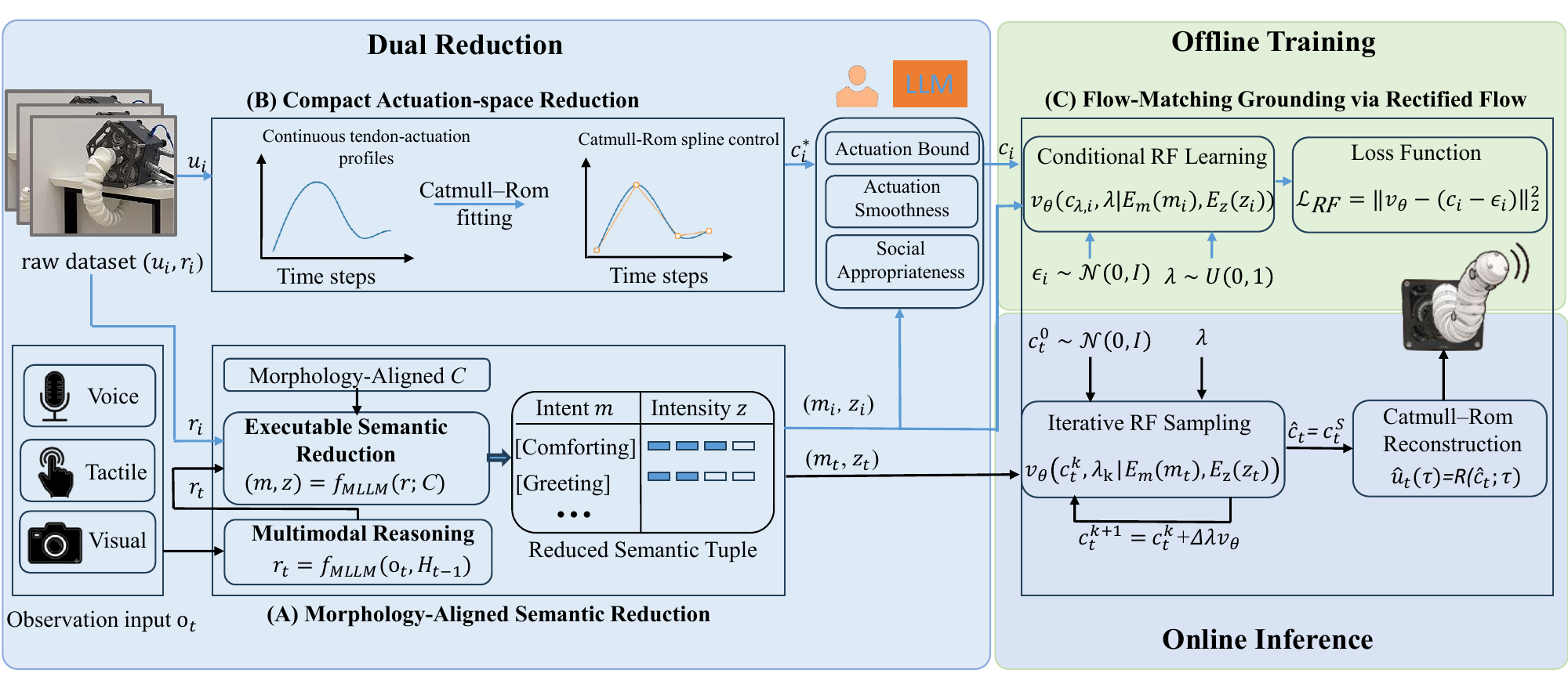}
   \caption{Details of three proposed modules:
morphology-aligned semantic reduction, compact actuation-space
reduction, and flow-matching grounding over reduced controls.}
   \label{fig:framework}
\end{figure*}

As shown in Fig.~\ref{fig:framework}(B), this component parameterizes each continuous actuation trajectory $u_i$ to a compact Catmull--Rom spline-control
matrix that captures the smooth, low-dimensional structure of expressive
trunk motion.
\subsubsection{Actuation Trajectory Generation}
Here, every continuous actuation trajectory $u_i$ is designed from elephant-inspired kinematic primitives,
including bending, curling, extension, retraction, and contact-aware
deformation~\cite{dagenais2021elephant,li2024biomimetic,
zhang2023preprogrammable,wang2025spirobs},
using existing whole-body shape-control strategies~\cite{almanzor2023static,kasaei2025shapeaware}. Here, let $u_i(\tau)\in\mathbb{R}^{N_u}$ denote the actuation at normalized
execution time $\tau\in[0,1]$ where $N_u$ is the number of actuation
channels.

\subsubsection{Spline-Control Encoding}
To obtain a compact and
structured representation of $u_i(\tau)$ for further efficient training, 
Catmull--Rom spline is introduced to encode this continuous trajectory. A Catmull--Rom spline is a piecewise cubic interpolating curve whose
segments pass through the control points with local support and $C^1$
continuity~\cite{derose1988geometric}. Accordingly, each actuation
trajectory $u_i(\tau)$ is represented by $P$ control points collected
in a compact matrix $c_i\in\mathbb{R}^{N_u\times P}$. The resulting spline learns its control points, provides
local control, and is inexpensive for further training. 


The control matrix is obtained by solving
\begin{equation}\label{ci}
\small
c_i^*=\arg\min_{c\in\mathbb{R}^{N_u\times P}}
  \int_{0}^{1}
  \bigl\|R(c;\tau)-u_i(\tau)\bigr\|_2^2\,\mathrm{d}\tau.
\end{equation}
where $R(c;\tau)\in\mathbb{R}^{N_u}$ is the Catmull--Rom reconstruction at
execution time $\tau$. The learned control matrix $c_i^*$ serves as the reduced training target.

\subsection{Flow-Matching Grounding via Rectified Flow}
Fig.~\ref{fig:framework}(C) summarizes the offline training and
online sampling of the conditional flow-matching model over reduced
spline controls. 


\subsubsection{Reduced Demonstration Set Construction}
\label{subsec:C1}

The offline demonstration library is constructed using the same
response-reasoning and semantic-reduction procedures as those used
during online interaction in Sec.~\ref{subsec:reduction}. Specifically, the response-reasoning
process defined in Eq.~\eqref{eq:2} is instantiated over a collection
of offline multimodal cue--history instances to generate
open-vocabulary response descriptions $r_i$. The same semantic
reduction rule as in Eq.~\eqref{eq:4} is then applied to each response,
yielding $
(m_i,z_i)=f_{\mathrm{MLLM}}(r_i;C),$
where $(m_i,z_i)\in\mathcal{M}\times\mathcal{Z}$ serves as the
semantic condition for the corresponding demonstration family.
Consequently, the offline library is organized over the same supported
semantic space $\mathcal{M}\times\mathcal{Z}$ used during online
inference. 

For each condition $(m_i,z_i)$, candidate actuation trajectories are designed and then fitted into Catmull–Rom matrices $c_i^*$ using \eqref{ci}. Note that $c_i^*$ candidates sharing the same intent--intensity tuple may differ in bending
direction, reach or curvature, temporal profile, and interaction
proximity. The expert Catmull–Rom matrices $c_i$ are retained after
actuation-bound, actuation-smoothness, and social-appropriateness checks for $R(c_i^*;\tau)$ via fixed-prompt MLLM screening for
semantic agreement with $(m_i,z_i)$~\cite{zheng2023judging}, and manual verification, on the robotic platform.
Finally, the resulting expert demo set is
\begin{equation}
  \small \mathcal{D}_{\mathrm{demo}}
  =
  \{(c_i,m_i,z_i)\}_{i=1}^{N_{\mathrm{demo}}}.
\end{equation}



\subsubsection{Training via Rectified Flow}
Given the reduced training set $\mathcal{D}_{\mathrm{demo}}$, the
grounding model learns a conditional distribution over spline-control
matrices,
\begin{equation}
\small
    c \sim p_\theta\!\left(c\mid E_m(m), E_z(z)\right),
    \qquad
    c\in\mathbb{R}^{N_u\times P},
\end{equation}
where $E_m(\cdot)$ is a pretrained text encoder~\cite{radford2021learning}
that embeds the intent label, and \(E_z(\cdot)\) is a learnable
embedding for the discrete intensity level. 
Because each intent--intensity tuple represents a motion family rather
than a unique trajectory, deterministic regression may average distinct
valid realizations and suppress within-tuple variation. 
Therefore, in the context of real-time HRI grounding, a lightweight conditional flow-matching method, rectified flow~\cite{lipman2023flow,liu2022flow}, is adopted here. It directly learns a continuous
velocity field over the compact spline-control space, retaining
one-to-many generation while allowing inference through the
explicit Euler method, more efficiently than heavier diffusion methods~\cite{ho2020denoising}.
For each \((c_i,m_i,z_i)\in\mathcal{D}_{\mathrm{demo}}\), a noise
matrix \(\epsilon_i\sim\mathcal{N}(0,I)\) with the same shape as \(c_i\)
and a flow time \(\lambda\sim\mathcal{U}(0,1)\) are sampled. The interpolated control and its target velocity are defined as
\begin{equation}
\small
    c_{\lambda,i}
    =
    (1-\lambda)\epsilon_i+\lambda c_i.
\end{equation}
 The velocity field is regressed onto the straight-line target,
\begin{equation}
\label{eq:rf_loss}
    \small    \mathcal{L}_{\mathrm{RF}}
    = \mathbb{E}\!\left[
      \left\| v_\theta\!\left(c_{\lambda,i},\lambda\mid E_m(m_i),E_z(z_i)\right)
      - (c_i-\epsilon_i)\right\|_2^2
    \right],
\end{equation}

\subsubsection{Inference and Trajectory Reconstruction}

 At inference, the semantic tuple \((m_t,z_t)\) obtained from
Eq.~\eqref{eq:4} conditions the learned velocity field. Starting from \(c_t^0=\epsilon_t\),
\(\epsilon_t\sim\mathcal{N}(0,I)\), the learned flow is integrated
using \(S\) explicit-Euler steps:
\begin{equation}
\label{eq:rf_inference}
\small
\begin{aligned}
    c_t^{k+1}
    &=
    c_t^k
    +
    \Delta\lambda\,
    v_\theta\!\left(
      c_t^k,\lambda_k
      \mid E_m(m_t),E_z(z_t)
    \right),
\end{aligned}
\end{equation}
where $\lambda_k=\frac{k}{S},
    \ 
    \Delta\lambda=\frac{1}{S},
    \ 
    k=0,\ldots,S-1$.
The final reduced control is reconstructed into a continuous actuation
trajectory as
\begin{equation}
\label{eq:reconstruction}
\small
    \hat u_t(\tau)=R(\hat c_t;\tau),
    \qquad
    \tau\in[0,1].
\end{equation}
where $\hat c_t=c_t^S$. Here, \(R\) is the Catmull--Rom reconstruction operator defined in~\eqref{ci}. The variable \(\lambda\)
indexes flow integration, whereas \(\tau\) denotes normalized motion
execution time.

\section{Experimental Validation}
\label{sec:experiments}

We evaluate the proposed framework in two stages: a  
grounding ablation that tests the contributions of semantic reduction, 
spline actuation reduction, and RF generation, followed by a physical HRI evaluation on a soft-trunk platform.
\begin{table}[t]
\scriptsize
\centering
\caption{Experimental implementation settings.}
\label{tab:impl}
\setlength{\tabcolsep}{3.0pt}
\begin{tabular}{p{35mm}p{51mm}}
\toprule
\multicolumn{2}{l}{\textit{Morphology-aligned Semantic Reduction}} \\
\midrule
Base MLLM & Qwen-3-VL \\
Expert Tuple space & 16 intents; 4 discrete intensity level \\
\midrule
\multicolumn{2}{l}{\textit{Compact Actuation-space Reduction}} \\
\midrule
Number of Actuation Channels & \(N_u=4\) \\
Number of Time Discretization & \(T=150\) \\
Number of Catmull--Rom controls &  \(P=4\)
\\
\midrule
\multicolumn{2}{l}{\textit{Flow-Matching Grounding via Rectified Flow}} \\
\midrule
RF network & 4-layer MLP, width 256, SiLU; 128-d flow-time embedding \\
Optimization Parameters & AdamW, \(10^{-3}\); batch size 8; 10,000 steps \\
RF and DDPM Sampler & Euler solver, \(S=50\) steps \\
\midrule
\multicolumn{2}{l}{\textit{Semantic-to-Actuation Grounding Evaluation}} \\
\midrule
Demonstration set & \(N_{\mathrm{total}}=256\)\\
Held-out split &  \(N_{\mathrm{held-out}}=64\) \; \(N_{\mathrm{training}}=192\)  \\
Diversity evaluation & \(K_{\mathrm{div}}=30\) generated samples per held-out \\
Correctness tolerance & Normalized RMS threshold \(\tau_{\mathrm{rms}}=0.35\) \\
\midrule
\multicolumn{2}{l}{\textit{Embodied Evaluation}} \\
\midrule
Participants & 100 participants \\
Conditions & AV-only vs. grounded trunk motion \\

\bottomrule
\end{tabular}
\end{table}

\subsection{Ablation Experiment: Semantic-to-Actuation Grounding}
\label{subsec:main_grounding}

\begin{figure}[t]
    \centering
    \includegraphics[page=9, width=110mm,trim=60mm 45mm 30mm 55mm,clip]{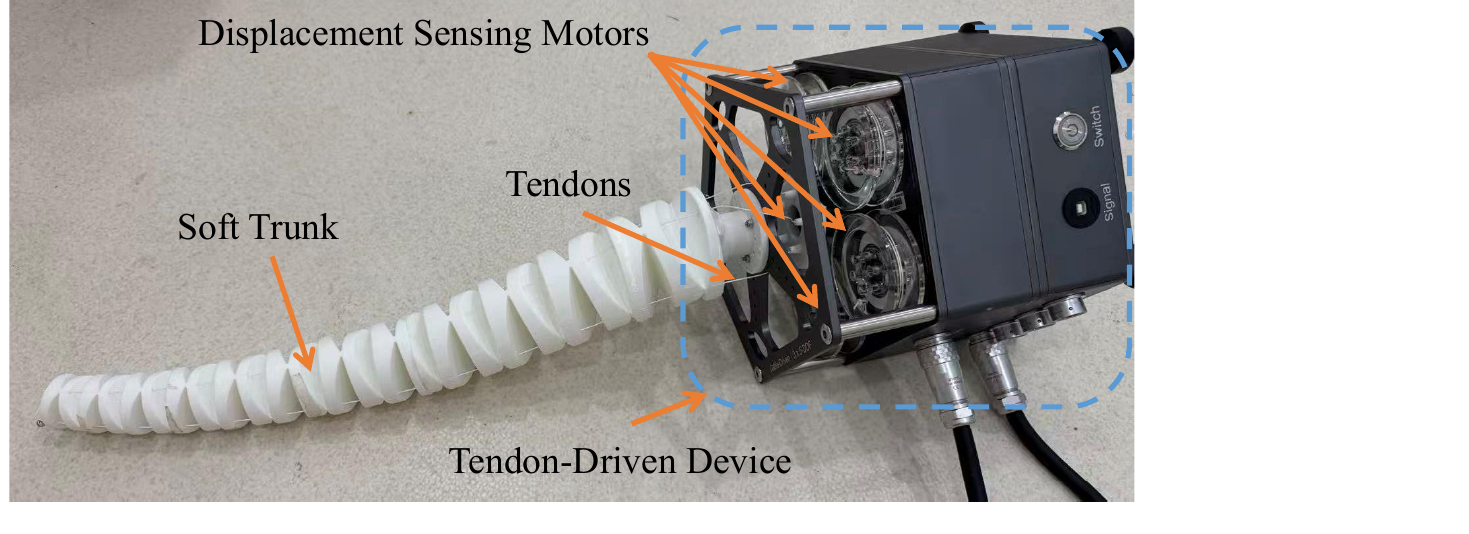}
    \caption{Tendon-driven soft-trunk platform.}
    \label{fig:setup}
\end{figure}

\subsubsection{Experimental Setup}

The implementation settings, including the hardware setup (Fig.~\ref{fig:setup}),
are summarized in Table~\ref{tab:impl}. 
The expert grounding dataset contains \(N_{\mathrm{total}}=256\) paired samples.
The retained demonstration library defined in Sec.~\ref{subsec:C1} is split
into \(N_{\mathrm{training}}=192\) training pairs and \(N_{\mathrm{held-out}}=64\) held-out pairs (library reference) in a tuple-balanced manner. The held-out set is denoted as
\begin{equation}\label{hd}
\small
\mathcal{D}_{\mathrm{held-out}}
=
\{(u^{\mathrm{tar}}_j,m_j,z_j)\}_{j=1}^{N_{\mathrm{held-out}}}.
\end{equation}
All methods are evaluated after their outputs are reconstructed or
represented in the common discrete actuation format specified in
Table~\ref{tab:impl}, consisting of a \(T=150\)-step trajectory over
\(N_u=4\) tendon-driven actuation channels.

\subsubsection{Compared Methods}

Each method is named following the pattern:
{\textbf{\emph{Semantic Condition + Actuation Representation + Generator}}}.

\textbf{\textit{Semantic Condition:}} \textit{Raw} means that the generator is conditioned directly on the
open-vocabulary MLLM response representation. \textit{Tuple}
means that the generator is conditioned on the morphology-aligned
intent--intensity tuple defined in Sec.~\ref{Morp}.

\textbf{\textit{Actuation Representation:}} \textit{Dense} means that the model learns and predicts the dense actuation trajectory $u^{\mathrm{d}}_i(j) \in \mathbb{R}^{N_u\times T}, j=1,2,3...T$ discretized from $u_i(\tau)\in\mathbb{R}^{N_u}$, as shown in Table~\ref{tab:impl}. \textit{Spline} means that the proposed actuation-space reduction method learns and predicts the reduced Catmull--Rom control matrix \(\hat{c}\in\mathbb{R}^{N_u\times P}\).

\textbf{\textit{Generator:}} \textit{CNN} denotes a deterministic supervised regressor. \textit{DDPM} denotes a denoising-diffusion
generator. \textit{RF} denotes the proposed rectified-flow generator.

\subsubsection{Metrics}

Three metrics are reported: held-out grounding correctness, valid-sample
diversity, and inference time. 

\textbf{Success rate.}
For each held-out intent--intensity tuple \((m_j,z_j)\), we use the
corresponding motion-library trajectory
\(u^{\mathrm{tar}}_j\) as the target actuation sequence.
The actuation error between the inferred trajectory
\(\hat{u}_j\) conditioned on the same $(m_j,z_j)$ from the held-out set \eqref{hd} and the target trajectory is measured by the normalized RMS distance:
\begin{equation}
\small
\mathrm{RMS}_j
=
\sqrt{
\frac{1}{N_uT}
\left\|\hat{u}_j-u^{\mathrm{tar}}_j\right\|_F^2
}.
\end{equation}
Lower RMS values indicate closer agreement between the generated and intended
target actuation trajectories. We count a generated trajectory as correct when its RMS is below a
fixed tolerance \(\tau_{\mathrm{rms}}\):
\begin{equation}
\small
\mathrm{Succ}_{\mathrm{held-out}}
=
\frac{1}{N_{\mathrm{held-out}}}
\sum_{j=1}^{N_{\mathrm{held-out}}}
\mathbb{I}
\left[
\mathrm{RMS}_j \leq \tau_{\mathrm{rms}}
\right].
\end{equation}{Several candidate RMS tolerances were calibrated against human
annotations and LLM-as-a-judge~\cite{zheng2023judging} assessments of the corresponding
hardware motions. The selected threshold
\(\tau_{\mathrm{rms}}=0.35\) showed the closest agreement between
RMS-based success and perceived motion validity. To verify that the
comparative conclusion is not an artifact of this tolerance, we further
evaluated
\(\tau_{\mathrm{rms}}\in\{0.25,0.30,0.35,0.40,0.45\}\). As expected,
absolute success rates increased with the tolerance, but the method
ordering remained unchanged across all tested thresholds. We therefore
report \(\tau_{\mathrm{rms}}=0.35\) as the annotation-calibrated operating
point.}

\begin{figure*}[t]
   \centering

   \includegraphics[page=7,width=178mm,trim=0mm 40mm 0mm 18mm,clip]{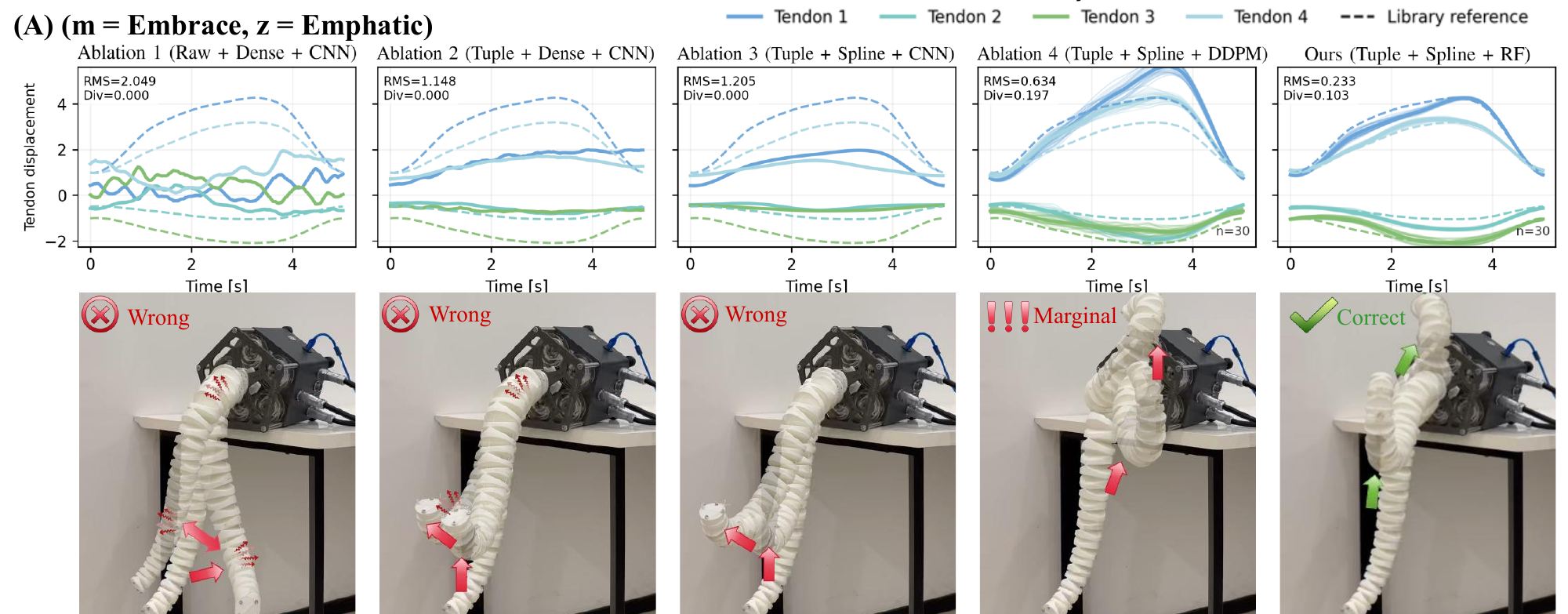}
   \vspace{-2mm}

   \includegraphics[page=8,width=178mm,trim=0mm 40mm 0mm 18mm,clip]{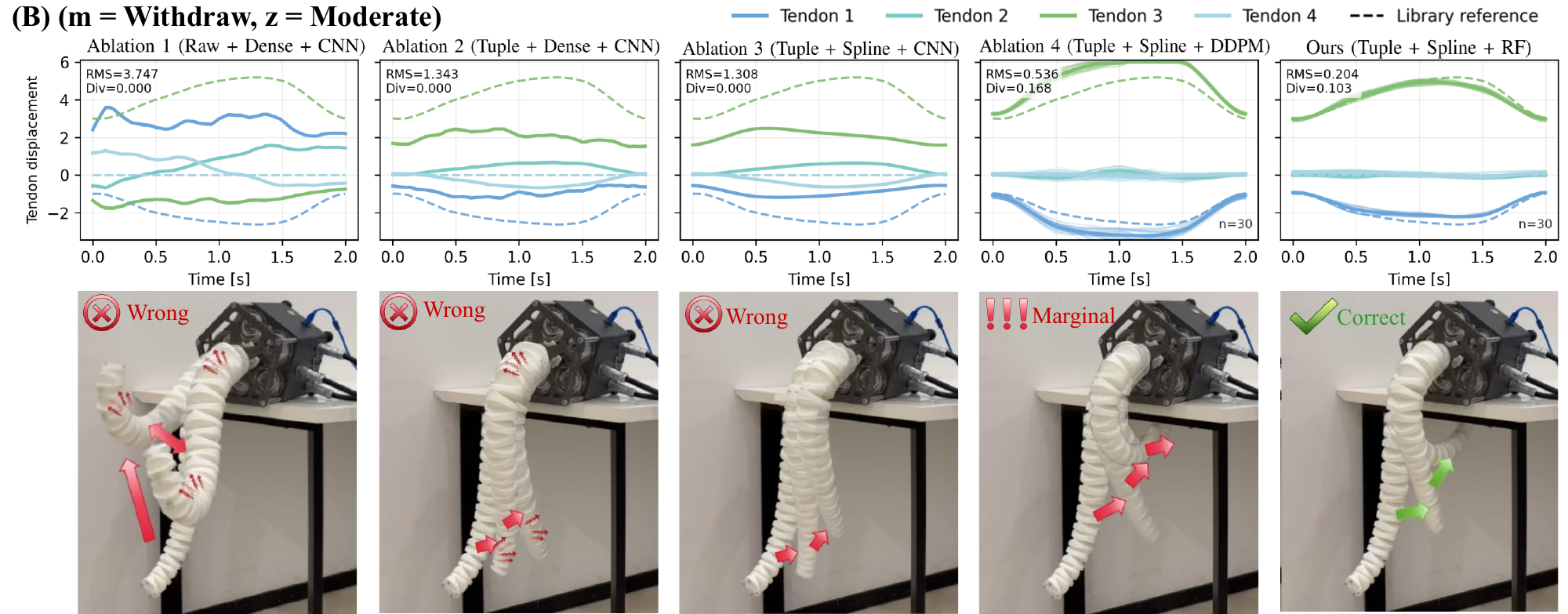}

   \caption{Held-out semantic-to-actuation grounding examples. The proposed Tuple + Spline + RF model more
closely follows the verified library reference while producing physically
consistent whole-body trunk motions.}
   \label{fig:example}
\end{figure*}

\textbf{Diversity.} For each held-out pair \((u_j^{tar},m_j,z_j)\), we infer $K_{div}$ trajectories
$\{\hat{u}_j^{(k)}\}_{k=1}^{K_{\mathrm{div}}}$. The diversity for each held-out sample is the mean pairwise RMS distance:
\begin{equation}
\small
D_j=
\frac{1}{\binom{K_{\mathrm{div}}}{2}}
\sum_{1\leq a<b\leq K_{\mathrm{div}}}
\mathrm{RMS}\!\left(
\hat{u}_j^{(a)},
\hat{u}_j^{(b)}
\right).
\end{equation}
The reported diversity is averaged over the complete held-out set:
\begin{equation}
\small
\mathrm{Div}
=
\frac{1}{|\mathcal{D}_{\mathrm{held\text{-}out}}|}
\sum_{(u_j^{\mathrm{tar}},m_j,z_j)
\in\mathcal{D}_{\mathrm{held\text{-}out}}}
D_j.
\end{equation}

\textbf{Inference time.}
For each held-out pair
$(u_j^{\mathrm{tar}},m_j,z_j)
\in\mathcal{D}_{\mathrm{held\text{-}out}}$,
the single-trajectory grounding latency $\Delta t_j$ is measured.
The reported inference time is averaged over the complete held-out set:
\begin{equation}
\small
\mathrm{Time}
=
\frac{1}{|\mathcal{D}_{\mathrm{held\text{-}out}}|}
\sum_{(u_j^{\mathrm{tar}},m_j,z_j)
\in\mathcal{D}_{\mathrm{held\text{-}out}}}
\Delta t_j .
\end{equation}

\subsubsection{Results}

\begin{table}[h]
\scriptsize
\centering
\caption{{Ablation semantic-to-actuation grounding results.}}
\label{tab:grounding_results}
\setlength{\tabcolsep}{3.0pt}
\begin{tabular}{lcccc}
\toprule
Methods & Succ. & Div. & Time (ms) \\
\midrule
Ablation 1 (Raw + Dense + CNN)       & 25.0\% & 0.000 & 0.85 \\
Ablation 2 (Tuple + Dense + CNN)     & 48.8\% & 0.000 & \textbf{0.51} \\
Ablation 3 (Tuple + Spline + CNN)    & 56.2\% & 0.000 & 1.89 \\
Ablation 4 (Tuple + Spline + DDPM)   & 71.9\% & \textbf{0.165} & 7.86 \\
Ours (Tuple + Spline + RF)     & \textbf{77.2\%} & 0.148 & 4.87\\
\bottomrule
\end{tabular}
\end{table}

Table~\ref{tab:grounding_results} and Fig.~\ref{fig:example} show consistent gains along the proposed grounding pipeline.
Ablation 1 achieves only 25.0\% held-out success. In the emphatic embrace case (Fig.~\ref{fig:example}(A)), the commands oscillate
irregularly.  In the
moderate withdraw case (Fig.~\ref{fig:example}(B)), its tendon commands drift
away from the dashed library reference, producing a mismatched posture. Adding morphology-aligned semantic reduction (ablation 2)
raises success to 48.8\%.
Although the overall trend becomes closer to the dashed library reference,
the dense prediction still produces jittery and insufficiently smooth
tendon commands, so the executed postures are still marked as Wrong in
Fig.~\ref{fig:example}.
Compact actuation-space reduction (ablation 3) further improves success to 56.2\%. Shown in Fig.~\ref{fig:example}, its tendon
commands are smoother and more temporally structured, but the
deterministic regressor (CNN) averages multiple valid realizations: the curves
remain separated from the dashed library reference and produce under-expressive
final postures in both cases. With dual reductions, distributional generators (ablation 4) reaches 71.9\% success and the highest diversity of 0.165.
However, this higher sample
dispersion is accompanied by occasional marginal cases and outliers. In Fig.~\ref{fig:example}, the withdraw example is marginally close but still deviates from the dashed library reference ($\mathrm{RMS}=0.536$), while the
embrace example approaches
the actuation upper bound ($\mathrm{RMS}=0.634$).



As shown in Fig.~\ref{fig:example}, the proposed framework follows the dashed library references more closely in both examples and produces the intended withdraw and embrace postures with RMS errors of 0.204 and 0.233, respectively, both below $\tau_{\mathrm{rms}}=0.35$. These results are consistent with the quantitative trends in Table~\ref{tab:grounding_results}. The deterministic CNN baselines fail to capture one-to-many grounding behavior and tend to average possible motions, leading to collapsed diversity. In contrast, the rectified-flow model learns a conditional distribution over spline controls, thereby retaining one-to-many generation while improving held-out grounding correctness. Compared with DDPM, rectified flow avoids the iterative reverse-denoising sampling process and provides a simpler and faster generation procedure. Overall, the complete Tuple + Spline + RF model achieves the highest held-out success rate of 77.2\%, preserves valid motion diversity with a diversity score of 0.148, and runs faster than DDPM, reducing inference time from 7.86~ms to 4.87~ms.

\subsection{Physical HRI Study}
\label{subsec:embodied_eval}

\begin{figure}[!]
    \centering
    \includegraphics[page=32,width=\linewidth,trim=5mm 5mm 15mm 5mm,clip]{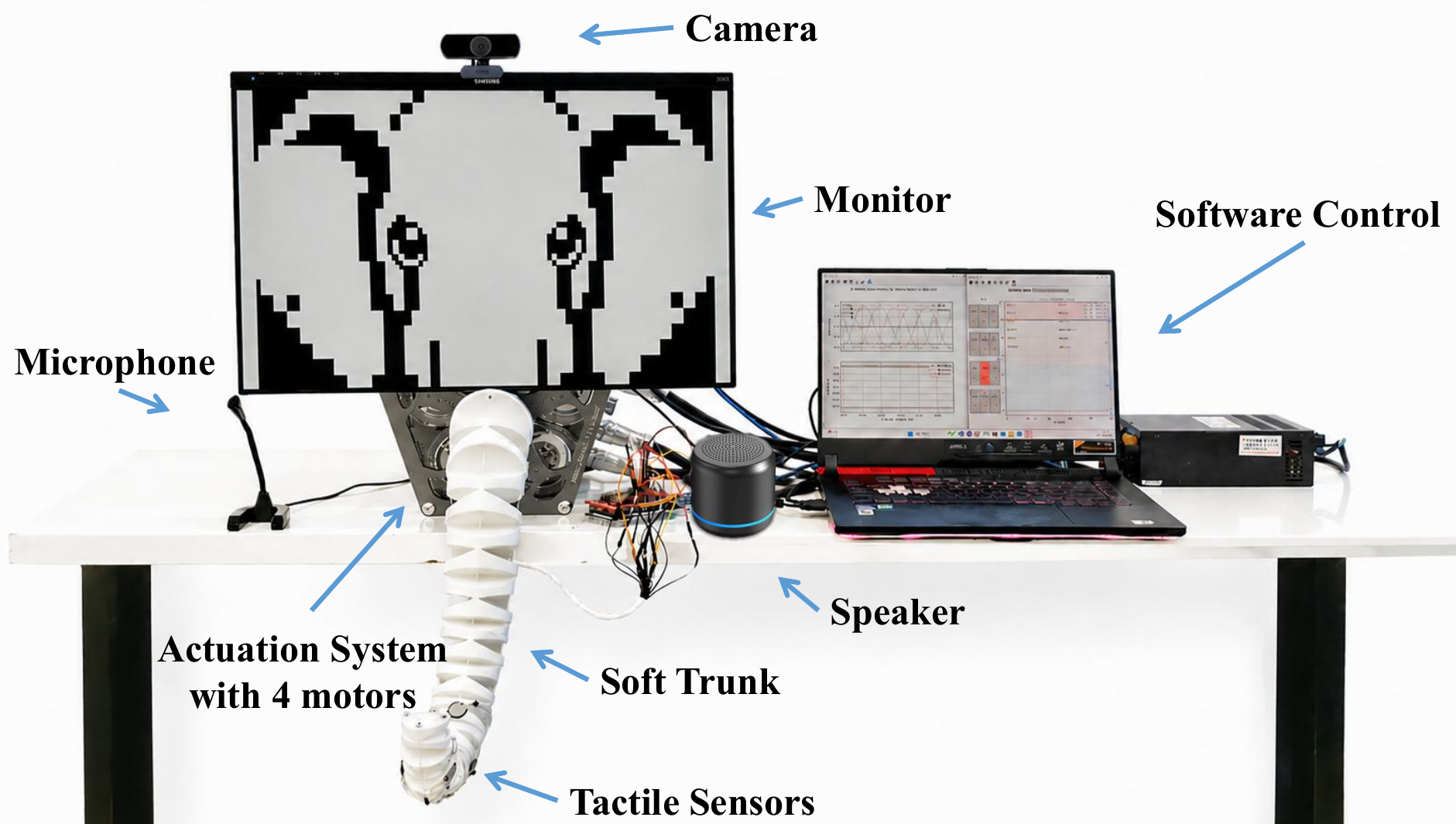}
    \caption{Elephant-inspired Robotic Interaction Companion (ERIC) used for physical HRI study.}
    \label{fig:eric_platform}
\end{figure}

This study evaluates whether adding generated soft-trunk motion improves human--robot interaction relative to the same audiovisual response. We instantiate the physical platform as
the Elephant-inspired Robotic Interaction Companion (ERIC) shown in
Fig.~\ref{fig:eric_platform}.  ERIC integrates a speech interface (speaker and mic), a facial-display monitor, a camera, tactile sensing, and a tendon-driven soft elephant continuum trunk for close-contact HRI.

\begin{figure*}[b]
  \centering      \includegraphics[page=10,width=178mm,trim=20mm 100mm 5mm 5mm,clip]{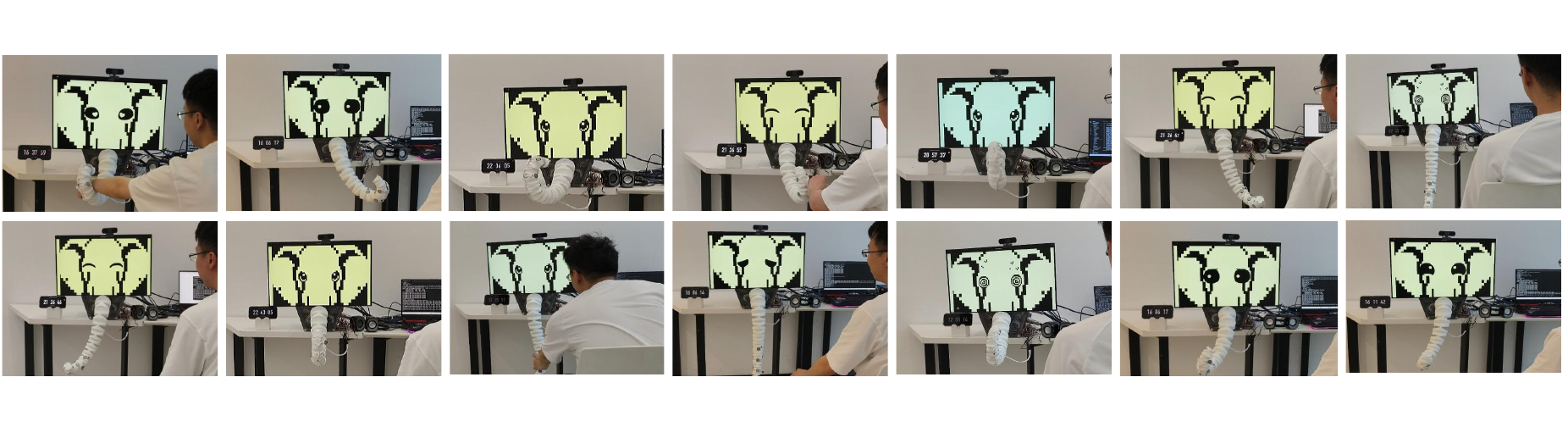}
   \caption{Representative interaction episodes between the user and ERIC.}
   \label{fig:example2}
\end{figure*}
\paragraph{Evaluation data}
We recruit 100 participants in a within-subject study, as stated in Table~\ref{tab:impl}. Each participant
interacts with ERIC under two counterbalanced conditions.

\paragraph{Baselines and evaluation metrics}
We compare two interaction conditions:
\begin{itemize}
\small
    \setlength{\itemsep}{0pt}
    \setlength{\topsep}{1pt}
    \item \textbf{AV-only}: ERIC responds with speech and facial display,
    while the soft trunk remains disabled.
    {
    \item \textbf{AV with physical trunk}: ERIC uses the
same audiovisual response, but additionally executes the
soft-trunk motion generated by the proposed semantic-to-actuation
grounding pipeline.}
\end{itemize}
Participants rate the interaction on a 5-point Likert
scale across six 
questionnaire\footnote{\href{https://docs.google.com/forms/d/e/1FAIpQLScjXhPsLinkvz92Q6Onjcx71oHfcTjkfmbkrfJOdump52KZOw/viewform?usp=sharing&ouid=111330114482771627245}{Google Form Questionnaire}} dimensions: naturalness, appropriateness, expressiveness, perceived safety, engagement, and overall satisfaction. We report the positive-rating ratio, defined as the fraction of ratings equal to 4 or 5. 

\paragraph{Results}
Representative interaction episodes between the user and ERIC are
shown in Fig.~\ref{fig:example2}. These examples qualitatively illustrate the
breadth of the physically realized behavior repertoire. Across different
interaction contexts, ERIC produces distinct whole-body trunk configurations
with variations in bending direction, curvature, reach, and interaction
proximity, while the generated motions remain smooth and executable on the
physical platform. Presented together with the audiovisual responses, the
trunk motions provide a visible embodied interaction channel.

\begin{figure}[t]
    \centering
    \includegraphics[page=19,width=90mm,trim=3mm 2mm 45mm 6mm,clip]{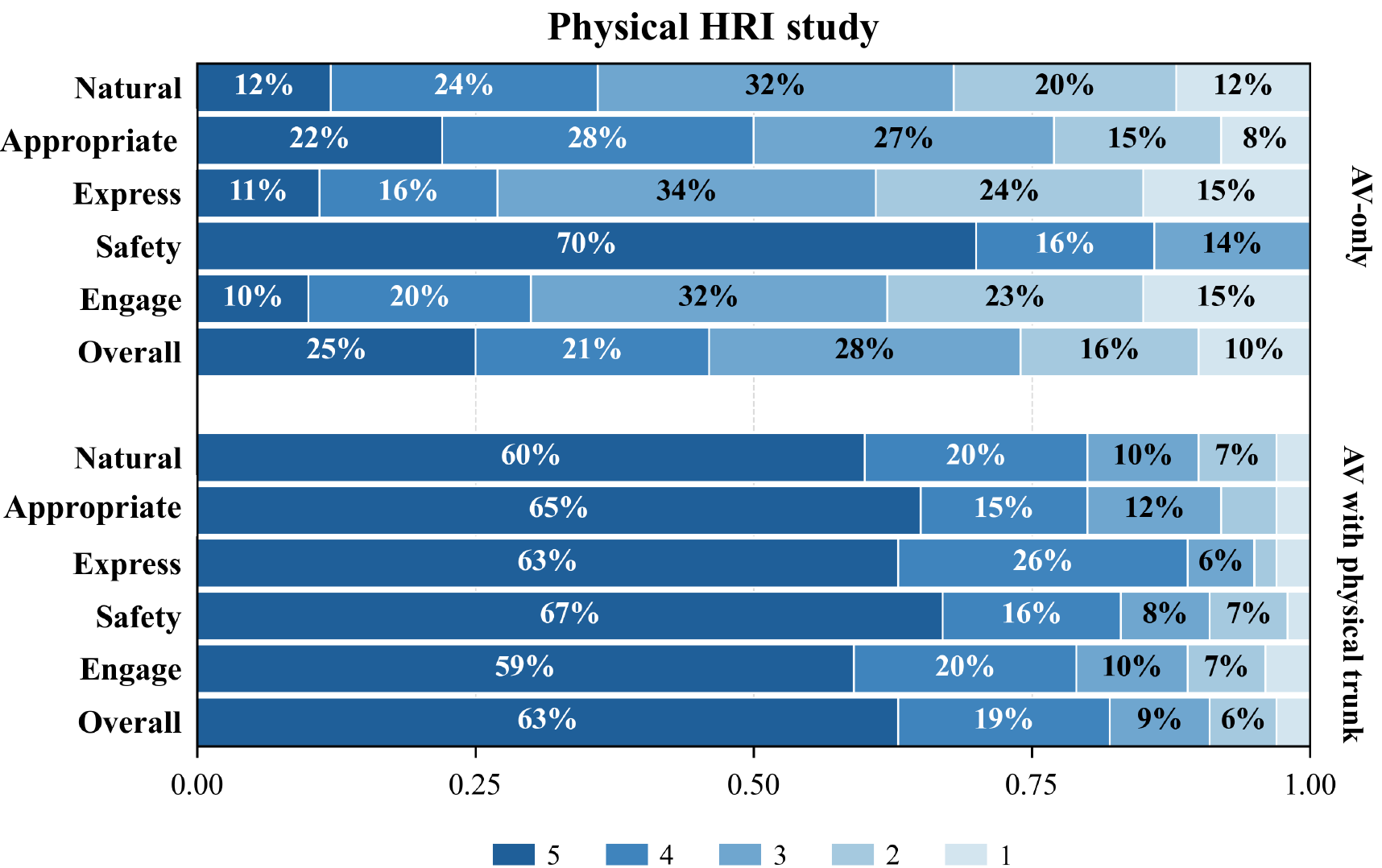}
    \caption{Physical HRI study with 100 participants.}
    \label{fig:questionnaire}
\end{figure}

The corresponding user-facing results are reported in
Fig.~\ref{fig:questionnaire}. Compared with the AV-only baseline, the
grounded-trunk condition increases the positive-rating ratio for naturalness
from 36\% to 80\% and for appropriateness from 50\% to 80\%, indicating that
the additional embodied-motion channel makes the overall response feel more
complete and better matched to the interaction context. The largest gains are
observed in expressiveness and engagement, which rise from 27\% to 89\% and
from 30\% to 79\%, respectively. Overall satisfaction also increases from
46\% to 82\%. {Perceived safety remains high in both conditions, with positive ratings of
86\% for AV-only and 83\% for grounded trunk motion. Although the grounded
condition introduces a visible moving appendage near the user, participants
continue to report a strong subjective sense of safety. Taken together, the qualitative examples in Fig.~\ref{fig:example2} and the
questionnaire results in Fig.~\ref{fig:questionnaire} indicate that the generated
grounded-trunk motions are physically realizable and
positively received as an embodied interaction channel. These results provide a user-facing validation that complements the ablation results in Sec.~\ref{subsec:main_grounding}.} Since the baseline is
audiovisual-only interaction, this study evaluates the added value of the
generated soft-trunk motion channel, while a direct perceptual comparison
with manually scripted trunk motions is left for future work.



\section{Conclusion}

In this work, we proposed a whole-body semantic-to-actuation grounding
framework for close-contact soft-trunk human--robot interaction. By
introducing morphology-aligned semantic reduction, compact actuation-space
reduction, and a lightweight flow-matching generator, the framework grounds
open-vocabulary semantic responses into executable trunk motions while
preserving motion diversity. Experiments show that the proposed approach
improves held-out grounding correctness from $25.0\%$ to $77.2\%$ and
achieves faster and more accurate generation than the diffusion baseline
($4.87$ ms vs. $7.86$ ms, $77.2\%$ vs. $71.9\%$). A physical HRI study with
100 participants further validates the deployability and user-facing benefit
of the generated soft-trunk motion channel, increasing the positive
overall-satisfaction rating from $46\%$ to $82\%$ over the audiovisual-only
baseline. Future work will investigate closed-loop semantic-to-actuation
grounding with real-time visual, tactile, and proprioceptive feedback,
toward adaptive whole-body soft-trunk responses during physical
human--robot interaction.

\bibliographystyle{IEEEtran}
\bibliography{reference}

\end{document}